\documentclass{article} % For LaTeX2e
\usepackage{iclr2025_conference,times}

\usepackage{amsmath,amsfonts,bm}

\def\eqref#1{equation~\ref{#1}}
\def\1{\bm{1}}

\def\rt{{\textnormal{t}}}

\def\rvx{{\mathbf{x}}}

\def\vb{{\bm{b}}}

\def\vm{{\bm{m}}}

\def\vw{{\bm{w}}}
\def\vx{{\bm{x}}}
\def\vy{{\bm{y}}}
\def\vz{{\bm{z}}}

\def\mE{{\bm{E}}}

\def\mI{{\bm{I}}}

\def\mV{{\bm{V}}}
\def\mW{{\bm{W}}}

\DeclareMathAlphabet{\mathsfit}{\encodingdefault}{\sfdefault}{m}{sl}
\SetMathAlphabet{\mathsfit}{bold}{\encodingdefault}{\sfdefault}{bx}{n}

\newcommand{\E}{\mathbb{E}}

\newcommand{\R}{\mathbb{R}}

\usepackage{amsthm}
\usepackage{booktabs}
\usepackage{hyperref}
\usepackage{url}
\usepackage{lipsum}
\usepackage{graphicx}
\usepackage{adjustbox}
\usepackage{wrapfig}
\usepackage{thmtools}

\newcommand{\tablestyle}[2]{\setlength{\tabcolsep}{#1}\renewcommand{\arraystretch}{#2}\centering\footnotesize}

\graphicspath{{./figures/}}

\newcommand{\fsl}{FLoWN}

\title{Flow to Learn: Flow Matching on Neural\\ Network Parameters}

\author{
\parbox{\linewidth}{
   Daniel Saragih\thanks{Main correspondence: \texttt{daniel.saragih@mail.utoronto.ca}} $^{, 1}$\,
    Deyu Cao$^{1, 2,\dagger}$ \,
    Tejas Balaji$^{1, \dagger}$\,
    Ashwin Santhosh$^{1}$\,
    }%
    \vspace{0.7em}\\
   $^{1}$University of Toronto \quad $^{2}$University of Tokyo
}

\iclrfinalcopy % Uncomment for camera-ready version, but NOT for submission.
\begin{document}

\maketitle
\def\thefootnote{$\dagger$}\footnotetext{Equal contribution}
\begin{abstract}
Foundational language models show a remarkable ability to learn new concepts during inference via context data. However, similar work for images lag behind. To address this challenge, we introduce \fsl{}, a flow matching model that learns to generate neural network parameters for different tasks. Our approach models the flow on latent space, while conditioning the process on context data. Experiments verify that \fsl{} attains various desiderata for a meta-learning model. In addition, it matches or exceeds baselines on in-distribution tasks, provides better initializations for classifier training, and is performant on out-of-distribution few-shot tasks while having a fine-tuning mechanism to improve performance.
\end{abstract}

\section{Introduction}
\label{sec:intro}
Flow matching (FM) \citep{albergo2023building, lipmanFlowMatchingGenerative2023, liu2023flow} is a prominent fixture in generative modeling tasks from imaging \citep{lipmanFlowMatchingGenerative2023, tongImprovingGeneralizingFlowbased2024, esser2024scalingrectifiedflowtransformers, liu2024instaflow} to language \citep{gat2024discrete, shaul2024flowmatchinggeneraldiscrete, campbell2024generative}. However, its application to neural network weights has not been explored. By leveraging the principled, yet versatile training of FM, we aim to generate task-specific weights on novel tasks. 

Multiple approaches have been tried to generate weights capable of few-shot learning (FSL), motivated by its speed compared to conventional training. For instance, various diffusion-based approaches \citep{soroDiffusionbasedNeuralNetwork2024, zhang2024metadiff, wangNeuralNetworkDiffusion2024} have been used to generate neural network weights. However, flexibility is limited by its restriction to Gaussian processes and a sluggish inference speed.
More broadly, we may categorize this form of learning as meta-learning \citep{fiftyContextAwareMetaLearning2024, huPushingLimitsSimple2022, zhmoginovHyperTransformerModelGeneration2022a}, which aims to learn concepts from a few demonstrations. It is therefore natural that we have two evaluation settings: in-distribution tasks and out-of-distribution (OOD) tasks. With enough training and capacity, it's clear \textit{meta-models} (i.e. models trained on multiple data distributions) should excel at in-distribution tasks. However, generalization to novel tasks often presents a challenge to meta-learning and weight generation frameworks \citep{wangNeuralNetworkDiffusion2024, schuerholt2024sane}. Recently, \citet{soroDiffusionbasedNeuralNetwork2024} goes to some extent to cross this generalization gap, however, we show that some improvements can be made to the approach. See Appendix \ref{app:rel-works} for further related works.

In this paper, we introduce Flow-based Learning of Weights for Neural adaptation (\fsl{}), a new class of method for weight generation. Empirical evaluations validate the following contributions: \textbf{1)} The generated weights match or exceed conventionally trained models on in-distribution tasks, and provide better initializations for fine-tuning on OOD tasks, \textbf{2)} \fsl{} is able to conditionally retrieve pre-trained weights from a distribution pre-trained on various datasets while matching their performance, \textbf{3)} \fsl{} is capable of performing well on OOD few-shot tasks while having a fine-tuning mechanism to improve performance.

\section{Methods}
\label{sec:methods}
\subsection{Preliminaries}
\paragraph{Conditional flow models.} \citet{chenNeuralOrdinaryDifferential2019} first introduced continuous normalizing flows as an effective data generation process through modeling dynamics. Simulation-free methods improve on this concept by simplifying the training objective \citep{lipmanFlowMatchingGenerative2023, albergo2023building, liu2023flow}. Following the formulation of \citet{lipmanFlowMatchingGenerative2023}, given random variables $\bar \rvx_0 \sim p_0$ and $\bar \rvx_1 \sim p_1$ a data distribution, define a reference flow $\bar \rvx = (\bar \rvx_t)_{t \in [0, 1]}$ where $\bar \rvx_t = \beta_t \bar \rvx_0 + \alpha_t \bar \rvx_1$ with the constraint that $\alpha_0 = \beta_1 = 0$ and $\alpha_1 = \beta_0 = 1$. The aim of flow modeling is to learn a sequence $\rvx = (\rvx_t)_{t \in [0, 1]}$ which has the same marginal distribution as $\bar \rvx$. To make this a feasible task, we describe this process as an ODE: $d\rvx_t = v(\rvx_t, t) dt$ where $\rvx_0 \sim \mathcal{N} (0 , \mI)$. Training proceeds by first parameterizing $v(\rvx_t, t)$ by a neural network $\theta$ and matching the reference flow velocity, i.e. $u(\rvx_t, t) := \frac{d}{d t} \bar \rvx_t$. This would, however, be an unfeasible training objective, therefore, we condition on samples from the terminal distribution $\rvx_1 \sim p_1$ and train
\begin{equation}
    \label{eq:cfm}
    L_{\text{cfm}}(\theta) = \E_{\rt \sim U[0, 1], \rvx_1 \sim p_1, \rvx_t \sim p_t(\cdot | \rvx_1)}||v_\theta(\rvx_t, t) - u(\rvx_t, t, \rvx_1)||.
\end{equation}
\citet{lipmanFlowMatchingGenerative2023} proved that this loss produces the same gradients as the marginal loss, thus optimizing it will result in convergence to the reference $u(\rvx_t, t)$. Moreover, we can always marginalize an independent conditioning variable $\vy$ on $v_\theta, u$ -- this will serve as our context conditioning vector.

% \paragraph{Flow matching with data-dependent potentials.} The straight line paths advocated in the conventional flow matching framework \citep{lipmanFlowMatchingGenerative2023} can be contextualized in the broader framework of matching interpolants $(\rvx_t)_{t \in [0, 1]}$ to minimize a given energy function. In the conventional case, the kinetic energy $\gE(\rvx_t, \dot \rvx_t) = \E_{\rt \sim U[0, 1]} ||{\dot \rvx_t}||^2$ is minimized, however, more general energies have been considered \citep{neklyudov2023actionmatching, neklyudov2024wassersteinlagrangian, liu2024generalized}. Of interest to us is the work on Metric Flow Matching \citep{kapusniakMetricFlowMatching2024} which seeks interpolants that minimize 
% \begin{equation}
%     \label{eq:mfm-loss}
%     \E_{(\rvx_0, \rvx_1) \sim \pi} \gE_g(\rvx_t) = \E_{\rt \sim U[0, 1], (\rvx_0, \rvx_1) \sim \pi} ||\dot \rvx_t||_{g(\rvx_t)}^2,
% \end{equation}
% where $\pi$ is some distribution on the product, typically $p_0 \otimes p_1$, and $g$ is a data-dependent Riemannian metric on the ambient space $\R^d$. \citet{kapusniakMetricFlowMatching2024} showed that the norm in \eqref{eq:mfm-loss} can be re-written as $||\dot \rvx_t||^2 + V_t(\rvx_t, \rvx_0, \rvx_1)$ where data-dependence is deferred to the potential $V_t(\cdot)$.

\paragraph{Few-shot learning.} The problem of few-shot learning is often formulated as a $n$-way-$k$-shot classification task. In particular, given $n$ classes and $k$ examples for each class, $\mathcal{S} = \{(\vx_i, \vy_i)\}_{i = 1}^{nk}$ of (image, label) \textit{support} set pairs, our meta-learner is tasked with classifying \textit{query} set images $\mathcal{Q} = \{\vx_{nk + 1}, \ldots, \vx_{nk + q}\}$. Relevant to our approach, recently, \citet{fiftyContextAwareMetaLearning2024} proposed an image meta-learning architecture, CAML, consisting of three components: a frozen pre-trained image encoder, a class encoder, and a transformer-based sequence model. ELMES, the class encoder, was shown to possess two attractive properties: label symmetry and permutation invariance. The transformer sequence model takes the concatenation of the image and label embeddings of the support set images, and a special "unknown" label embedding is used on query set images. These query images are analagous to the [CLS] tokens in transformers as the logits corresponding to the query images are then passed into a classifier MLP to predict labels.

\subsection{Flow to Learn}
We describe the components of our approach below and leave more details to Appendix \ref{app:arch}, \ref{app:training}.

\paragraph{Weight encoder.} Due to the intractable size of weight space, it is necessary for modeling to take place in latent space. We justify this design by appealing to work on the Lottery Ticket Hypothesis \citep{frankle2018the, liu2024surveylotterytickethypothesis} as well as the body of work on pruning \citep{cheng2024asurveyonpruning}, which suggests that, like natural data, neural networks live on a low-dimensional manifold within its ambient space. We use three encoder variants in this paper, first is a variational autoencoder (VAE) \citep{kingma2022autoencodingvariationalbayes} set up as in \citet{soroDiffusionbasedNeuralNetwork2024}, and the second is the graph-based encoder (GE) of \citet{kofinasGraphNeuralNetworks2024} which takes into account permutation invariance of neural networks, and third is based on the Learning on LoRAs (LoL) architecture \citep{putterman2024learningloras}. Usage of each variant is dependent on experimental settings.

\paragraph{Flow meta-model.} The backbone of our meta-learning framework is a conditional FM model following \citet{tongImprovingGeneralizingFlowbased2024}. We make use of the flexibility of FM to use a non-Gaussian prior, specifically the Kaiming uniform or normal initializations \citep{he2015delvingdeep}, as the source $p_0$. The data distribution $p_1$ of base model weights is experiment-dependent, however, broadly they are obtained by conventional training methods or through a model zoo \citep{schurholtModelZoosDataset2022}.

\paragraph{Conditioning model.} To condition our flow meta-model, we use a pre-trained CAML \citep{fiftyContextAwareMetaLearning2024} adapated by Low-rank adaptation modules (LoRAs) \citep{hu2022lora}. Our choice is due to the extensive training of the CAML architecture on several datasets as well as the principled label encoding used in their approach. As we are only interested in an overall encoding of support images, we append a combining layer to aggregate the image latents. The conditioning vector is incorporated by concatenating to the latent vector.

% \paragraph{Data-dependent energy.} Following the discussion of the previous section, we can make a choice of potential energy that is feasible to compute, and lead to suitable reference paths. In our case, we implement two approaches: first, that of \citet{kapusniakMetricFlowMatching2024} to compute a data-dependent metric $\mG(\theta; \gD)$ and hence a geodesic. Second, we do multi-marginal flow matching with intermediate weights grouped into discrete time-intervals. The latter requires a relaxation of the smoothness assumption, see App. \ref{app:metric-learning}.

\vspace{-1.0em}
\begin{table}[h]
\caption{Best validation accuracy of unconditional \fsl{} generation. \textit{orig} denotes base models trained conventionally and \textit{p-diff} those generated using p-diff \citep{wangNeuralNetworkDiffusion2024}.}
\label{tab:uncond}
\tablestyle{3pt}{1.0}
\begin{center}
\begin{adjustbox}{max width=\textwidth}
\begin{tabular}{l @{\hspace{20pt}}ccc ccc ccc ccc}
\toprule
&  \multicolumn{3}{c}{CIFAR100} & \multicolumn{3}{c}{CIFAR10} & \multicolumn{3}{c}{MNIST} & \multicolumn{3}{c}{STL10} \\
\cmidrule(lr){2-4} \cmidrule(lr){5-7} \cmidrule(lr){8-10} \cmidrule(lr){11-13}
Base Models  & orig. & \fsl{} & p-diff. & orig. & \fsl{} & p-diff. & orig. & \fsl{} & p-diff & orig. & \fsl{} & p-diff\\
\midrule
Resnet-18         & 71.45 & 71.42 & 71.40 & 94.54 & 94.36 & 94.36 & 99.68 & 99.65 & 99.65 & 62.00 & 62.00 & 62.24 \\
ViT-base          & 85.95 & 85.86 & 85.85 & 98.20 & 98.11 & 98.12 & 99.41 & 99.38 & 99.36 & 96.15 & 95.77 & 95.80 \\
ConvNext-tiny     & 85.06 & 85.12 & 85.17 & 98.03 & 97.89 & 97.90 & 99.42 & 99.41  & 99.40 & 95.95 & 95.63 & 95.63  \\
\midrule
CNN w/ VAE      & 32.09 & 31.36 & 30.67 & 72.53 & 69.97 & 69.18 & 98.93 & 98.91 & 98.93 & 53.88 & 53.50 & 53.86 \\
CNN w/ GE     & 32.09 & 31.73 & 31.81 & 72.53 & 72.15 & 72.09 & 98.93 & 98.89 & 98.89 & 53.88 & 53.64 & 53.80\\
\bottomrule
\end{tabular}
\end{adjustbox}
\vspace{-1em}
\end{center}
\end{table}

\begin{wraptable}{r}{.45\textwidth}
\vspace{-2.2em}
\caption{Mean validation accuracy of top-5 \fsl{} model retrievals, one with a mini-Imagenet prior.}
\vspace{-1em}
\label{tab:model-retrieval}
\tablestyle{2.0pt}{1.1}
\begin{center}
\begin{adjustbox}{max width=.45\textwidth}
\begin{tabular}{lcccc}
\toprule
Method  &  \multicolumn{1}{c}{{MNIST}} & \multicolumn{1}{c}{{F-MNIST}} & \multicolumn{1}{c}{{CIFAR10}} & \multicolumn{1}{c}{{STL10}}\\
\midrule
Original & 91.1 & 72.7 & 48.7 & 39.0 \\
\fsl{} w/ mIN-prior & 63.0 & 41.9 &  22.6 & 18.8\\
\fsl{} & 91.7 & 73.8 & 50.3 & 40.8\\
\bottomrule
\end{tabular}
\end{adjustbox}
\vspace{-3.5em}
\end{center}
\end{wraptable}

\section{Experiments}
\label{sec:experiments}
First, we confirm various properties that are to be expected of weight generation models. Next, we examine \fsl{}'s performance in few-shot learning. Further details are provided in~\ref{app:experimental-details}.

\subsection{Basic Properties of \fsl{}}

\paragraph{Unconditional generation.} We first evaluate the basic modeling capabilities of the flow meta-model. The target distribution $p_1$ is generated by training a variety of base models on known datasets: CIFAR10, CIFAR100, and MNIST, and saving 200 weight checkpoints each. For large models, we can choose to generate only a subset of the weights. In our case, we generate the batch norm parameters for ResNet-18 \citep{He2015DeepRL}, ViT-base \citep{dosovitskiy2021animage} and ConvNext-tiny \citep{liu2022convnet2020s}, and the full medium-CNN~\citep{schurholtModelZoosDataset2022}. The aim of this test is to train a separate meta-model for each dataset and validate its base model reconstruction on classifying its corresponding test set. Table \ref{tab:uncond} shows that we are able to match base models trained conventionally and with p-diff \citep{wangNeuralNetworkDiffusion2024}. 

\vspace{-1em}
\begin{table*}[h]
\caption{Mean validation accuracy of fine-tuned generated weights post-retrieval. The asterisk (*) indicates datasets on which the model was not trained.}
\label{tab:ft-retrieval}
\vspace{-1em}
{\fontsize{10}{10}\selectfont
\tablestyle{2.2pt}{1.0}
\renewcommand{\arraystretch}{0.9}
\setlength{\tabcolsep}{4pt}
\begin{center}

\begin{adjustbox}{max width=\textwidth}
    \begin{tabular}{l l c c c c c c c}
    \toprule
    Epoch & Method & MNIST & F-MNIST & CIFAR10 & STL10 & USPS* & SVHN* & KMNIST*\\
    \midrule
    \raisebox{-0.9\normalbaselineskip}[0pt][0pt]{0}
     & RandomInit & $\sim 10\%$ & $\sim 10\%$ & $\sim 10\%$ & $\sim 10\%$ & $\sim 10\%$ & $\sim 10\%$ & $\sim 10\%$ \\
     & \fsl{} & $83.58 \pm 0.58$ & $68.50\pm0.64$ & $45.93 \pm 0.57$ & $35.16 \pm 1.24$ &  $57.53\pm 2.43$& $17.99 \pm 0.82$& $11.79 \pm 0.51$\\
    \midrule
    \raisebox{-0.9\normalbaselineskip}[0pt][0pt]{1}
     & RandomInit & $18.12\pm 1.58$ & $26.90\pm 0.52$ & $28.75 \pm 0.22$ & $18.94 \pm 0.09$ & $17.69\pm 0.00$& $19.50\pm 0.03$& $14.48\pm 0.06$\\
     & \fsl{} & $84.49\pm0.65$ & $69.09\pm0.40$ & $46.85 \pm 0.30$ & $36.15 \pm 1.14$ & $72.45\pm 1.81$& $68.64 \pm 7.07$& $51.15 \pm 8.90$\\
    \midrule
    \raisebox{-0.9\normalbaselineskip}[0pt][0pt]{5}
     & RandomInit & $35.05\pm3.87$ & $51.08\pm2.15$ & $40.00 \pm 0.20$ & $28.24 \pm 0.01$ & $32.77\pm 0.46$& $39.59\pm 10.0$& $30.00\pm 0.30$\\
     & \fsl{} & $87.68\pm0.44$ & $70.32\pm0.50$ & $47.44 \pm 0.55$ & $37.43 \pm 1.19$ & $76.96\pm 1.29$& $77.36 \pm 1.07$& $69.14 \pm 10.1$\\
    \midrule
    \raisebox{-0.9\normalbaselineskip}[0pt][0pt]{25}
     & RandomInit & $87.70\pm0.90$ & $70.69\pm0.46$ & $46.86 \pm 0.01$ & $36.75 \pm 0.10$ & $82.02\pm 0.12$& $58.56\pm 19.5$& $55.05\pm 0.06$\\
     & \fsl{} & $92.29\pm0.41$ & $73.72\pm0.68$ & $49.25 \pm 0.73$ & $40.14 \pm 1.07$ & $82.28\pm 1.40$& $78.75 \pm 1.30$& $79.11 \pm 6.65$\\
    \midrule
    50
     & RandomInit & $92.76\pm0.08$ & $72.88\pm0.46$ & $48.85 \pm 0.74$ & $40.47 \pm 0.18$ & $88.35\pm 0.18$& $63.70\pm 22.1$& $64.32\pm 0.25$\\
    \bottomrule
    \end{tabular}
\end{adjustbox}
\end{center}
}
\vspace{-1em}
\end{table*}

\paragraph{Model retrieval and in-distribution initialization.} Following \citep{soroDiffusionbasedNeuralNetwork2024}, we perform model retrieval to test whether the meta-model can distinguish weights of the base model given conditioning samples from the dataset the base model was trained on. The base model is a simple 4-layer ConvNet and we obtain 100 weight checkpoints from the model zoo \citep{schurholtModelZoosDataset2022} for each dataset: MNIST, Fashion-MNIST (F-MNIST), CIFAR10, and STL10 after 46-50 epochs of conventional training. Unlike in the previous test, we will train just a single meta-model on 400 total base models conditioned on support samples from their training set via CAML. During validation, we pass in a random support sample from one of the four datasets and generate the \textit{full} ConvNet. In Table \ref{tab:model-retrieval}, we see that our top-5 validation accuracy matches that of the base models. Additionally, we repeated this experiment using weights from mini-Imagenet as a prior, but they seem to perform considerably worse than just Kaiming normal (see~\ref{app:model-retrieval} for a discussion). Next, we repeat this experiment but instead using weight checkpoints from epochs 21-25, and use the generated weights as an initialization before fine-tuning another 25 epochs. As shown in Table \ref{tab:ft-retrieval}, our initialization enjoys faster convergence, even for datasets on which the model was not trained, highlighting the generalization capability of our meta-model.

\begin{wraptable}{r}{0.5\textwidth}
\vspace{-2em}
\caption{Fine-tuning on OOD data. Data-F are results generated from fine-tuned meta-models, whereas Data-S are from static meta-models. Here, we generate the full CNN weights, whereas we only modify the batch norms of ResNet-18.}
\label{tab:ft}
\tablestyle{2.5pt}{1.0}
\begin{center}
\begin{adjustbox}{max width=0.5\textwidth}
\begin{tabular}{l @{\hspace{10pt}} cc cc cc}
\toprule
 &      \multicolumn{2}{c}{ResNet-18}  & \multicolumn{2}{c}{CNN w/ VAE} & \multicolumn{2}{c}{CNN w/ GE} \\
 \cmidrule(lr){2-3} \cmidrule(lr){4-5} \cmidrule(lr){6-7}
Base dataset & CIFAR10 & STL10 & CIFAR10 & STL10 & CIFAR10 & STL10\\
\midrule
CIFAR10-S & -- & 64.20 & -- & 24.01 & -- & 23.03 \\
CIFAR10-F & -- & 72.87 & -- & 60.09 & -- & 60.85\\
\midrule 
STL10-S & 93.09 & -- & 19.97 & -- & 18.13 & --\\
STL10-F & 94.06 & -- & 61.38 & -- & 69.42 & --\\
\bottomrule
\end{tabular}
\end{adjustbox}
\end{center}
\vspace{-1.5em}
\end{wraptable}

\paragraph{Fine-tuning the meta-model on OOD data.}\label{sec:model_retrieval} The setting of unconditional generation is quite restrictive as it is assumed that the output classifier has the same architecture and is to be used on the same dataset. In this experiment, we evaluate whether the meta-model can be effectively fine-tuned to achieve better performance on out-of-distribution data. We start with the meta-model trained from unconditional generation and generate weights for a different dataset. Subsequently, we compute the cross-entropy loss and backpropagate the gradients through the FM model. As this entails backpropagation through an ODE solver, we implement a stopgrad mechanism that restricts gradient flow before a time $0 < t' < 1$ to trade off accuracy for efficiency. Due to time constraints, we restrict ourselves to batch norms of ResNet-18 and the small-CNN. Table \ref{tab:ft} shows considerable improvement over generations obtained from a static FM meta-model and the VAE.

\subsection{Few-shot learning}
For this evaluation, we use mini-Imagenet \citep{vinyals2016matching} and a pre-trained ResNet-12 \citep{chen2021meta}. Following the typical FSL setting, the dataset is partitioned into meta-train and meta-test sets, and further into tasks whose size depends on the way and shot parameters. For instance, for 5-way-1-shot, the support set consists of one image from 5 classes, whereas the query set is always 15 images for each of the 5 classes in the support set. The in-distribution test entails labeling query images from the same dataset (i.e. trained on mini-Imagenet and evaluated on mini-Imagenet), whereas OOD tasks entail labeling novel query images. First, we train the ResNet-12 on the train split of mini-Imagenet; our goal is thus to generate a classifier head for each task.

\begin{wraptable}{r}{0.5\textwidth}
\vspace{-1.8em}
\caption{Few-shot learning accuracy on out-of-distribution tasks. We compare with D2NWG and best reconstruction of the classifier weights by our out-of-distribution VAE.}
\label{tab:fsl-small}
\vspace{-0.8em}
\begin{center}
\resizebox{0.5\columnwidth}{!}{
\begin{tabular}{lcc}
\toprule
Model & CIFAR10 & STL10\\
\midrule
Max acc. VAE Recon. & \textbf{73.1} & \textbf{80.4}\\
D2NWG \citep{soroDiffusionbasedNeuralNetwork2024} & 33.04 $\pm$ 0.04 & 50.42 $\pm$ 0.13 \\
\fsl{} & 51.46 $\pm$ 8.02 & 56.24 $\pm$ 8.96 \\
\bottomrule
\end{tabular}
}
\end{center}
\vspace{-0.8em}
\end{wraptable}
% In our case, we set 50,000 tasks in the meta-train set and 100 in the meta-test set. We perform this test by composing the target distribution using pre-trained weights from CLIP \citep{Radford2021LearningTV}, linear-probing a classifier head on top of the CLIP model for each of the 50,000 subsets for 100 epochs using AdamW optimizer with a learning rate of 1e-3 and weight decay of 1e-2.
In our case, we set 50,000 tasks in the meta-train set and 100 in the meta-test set. We perform this test by constructing a target distribution using pre-trained weights from ResNet-12, linear-probing a classifier head on top of the ResNet backbone for each of the 50,000 subsets for 100 epochs using the AdamW optimizer with a learning rate of $10^{-3}$ and weight decay of $10^{-2}$. Given our computational constraints, we evaluated \fsl{} on just two out-of-distribution datasets: CIFAR10 and STL10 by sampling weights 50 different times and taking the average top-3 accuracies. Table~\ref{tab:fsl-small} shows that our method achieves marginal gains on CIFAR10, but performance on STL10 remains below the baseline. Considering the high validation accuracies of our VAE, we anticipate that further training and tinkering, which we plan to conduct in subsequent revisions, will enhance \fsl{} generalization across tasks.

\section{Discussion and Future Work}
\label{sec:discussions}
In this work, we have provided a preliminary investigation of \fsl{} for weight generation with an application to few-shot learning. Future research directions include: \textbf{1)} training \fsl{} on a more comprehensive image dataset to improve efficacy on OOD tasks, \textbf{2)} a post-hoc fine-tuning mechanism \citep{domingo-enrichAdjointMatchingFinetuning2024} for adapting \fsl{} to difficult domains (e.g. medical imaging), \textbf{3)} incorporating intermediate base model weights obtained during conventional training to guide the inference trajectory of generated weights (e.g. via MetricFM \citep{kapusniakMetricFlowMatching2024}).

\subsubsection*{Acknowledgments}
The authors would like to thank Lazar Atanackovic and Kirill Neklyudov for helpful discussions in the planning stages and later discussions of the manuscript. 

\bibliography{./iclr2025_conference}
\bibliographystyle{iclr2025_conference}

\appendix
\section{Appendix}
This appendix consist of details left out in the main text. First, we perform a more comprehensive review of the related literature with further discussion of our motivations. Next, we go over the various components of \fsl{} and expound on their implementation and training procedure.

% \subsection{Proofs}
% \label{app:proofs}

% \begin{proof}[Proof of Theorem \ref{thm:ce}]
%     \lipsum[1]
% \end{proof}

\subsection{Related Works}
\label{app:rel-works}

\paragraph{Conditional flow matching.}
The CFM objective, where a conditional vector field is regressed to learn probability paths from a source to target distribution, was first introduced in \citet{lipmanFlowMatchingGenerative2023}. The CFM objective attempts to minimize the expected squared loss of a target conditional vector field (which is conditioned on training data and generates a desired probability path) and an unconditional neural network. The authors showed that optimizing the CFM objective is equivalent to optimizing the unconditional FM objective. Moreover, the further work~\citep{tongImprovingGeneralizingFlowbased2024} highlighted that certain choices of parameters for the probability paths led to the optimal conditional flow being equivalent to the optimal transport path between the initial and target data distributions, thus resulting in shorter inference times.  However, the original formulations of flow matching assumed that the initial distributions were Gaussian. \citet{pooladian2023multisample} extended the theory to arbitrary source distributions using minibatch sampling and proved a bound on the variance of the gradient of the objective. \citet{tongImprovingGeneralizingFlowbased2024} showed that using the 2-Wasserstein optimal transport map as the joint probability distribution of the initial and target data along with straight conditional probability paths results in a marginal vector field that solves the dynamical optimal transport problem between the initial and target distributions.

\paragraph{Neural network parameter generation.}
Due to the flexibility of neural network as function approximators, it is natural to think that they could be applied to neural network weights. \citet{denilPredictingParametersDeep2014} paved the way for this exploration as their work provided evidence of the redundancy of most network parameterizations, hence showing that paramter generation is a feasible objective.
Later, \citet{ha2017hypernetworks} introduced Hypernetworks which use embeddings of weights of neural network layers to generate new weights and apply their approach to dynamic weight generation of RNNs and LSTMs. A significant portion of our paper’s unconditional parameter generation section builds upon the ideas from \citet{wangNeuralNetworkDiffusion2024} and the concurrent work of \citet{soroDiffusionbasedNeuralNetwork2024} where the authors employ a latent diffusion model to generate new parameters for trained image classification networks. 

\paragraph{Meta-learning context.}
Although neural networks are adept at tasks on which they were trained, a common struggle of networks is generalization to unseen tasks. In contrast, humans can often learn new tasks when given only a few examples. A pioneering modern work in this field is MAML~\citep{finnModelAgnosticMetaLearningFast2017}, which learns good initialization parameters for the meta-learner such that it can easily be fine-tuned to new tasks. Their approach utilizes two nested training loops. The inner loop computes separate parameters adapted to each of the training tasks. The outer loop computes the loss using each of these parameters on their respective tasks and updates the model’s parameters through gradient descent. However, MAML often had unstable training runs, and so successive works gradually refined the method \citep{antoniou2018how, rajeswaran2019imaml, zhaoMetaLearningHypernetworks2020, Przewiezlikowski2022HyperMAMLFA}. The aforementioned works typically focus on classification tasks, however, this paradigm allows for great versatility. For instance, \citet{beckHypernetworksMetaReinforcementLearning2023} used hypernetworks to generate the parameters of a policy model and \citep{leeSupervisedPretrainingCan2023} exploited the in-context learning ability of transformers to general reinforcement learning tasks.

\paragraph{Weight generation for few-shot learning.}
Following up on the work of meta-learning context, few-shot learning is a natural application of such meta-learning algorithms. An early example is \citet{raviOptimizationModelFewShot2017} who designed a meta-learner based on the computations in an LSTM cell. At each training example in the support set, their meta learner uses the losses and the gradients of the losses of the base learner (in addition to other information from previous training examples) to produce base learner parameters for the next training example. The loss of the base learner on the test examples in the support set is backpropagated through the meta learner’s parameters. Moreover, we may leverage the advancements in generative modeling for weight generation. As we mentioned, \citet{leeSupervisedPretrainingCan2023} used transformers for in-context reinforcement learning, but we also see the works of \citet{zhmoginovHyperTransformerModelGeneration2022a, huPushingLimitsSimple2022,
kirsch2024generalpurposeincontextlearningmetalearning, fiftyContextAwareMetaLearning2024} use transformers and foundation models. More similar to our method is the body of work on using diffusion models for weight generation~\citep{duProtoDiffLearningLearn2023, zhang2024metadiff, wangNeuralNetworkDiffusion2024, soroDiffusionbasedNeuralNetwork2024}. These methods vary in their approach, some leveraging a relationship between the gradient descent algorithm and the denoising step in diffusion models to design their meta-learning algorithm. Others rely on the modeling capabilities of conditioned latent diffusion models to learn the target distribution of weights. Most evaluations conducted were in-distribution tasks, i.e. tasks sampled from the same data distribution as the training tasks, hence, there is room to explore ways of adapting this approach for out-of-distribution tasks.

% \subsection{Schr\"odinger bridge formulation of weight evolution}
% \label{app:sb-formulation}
% \lipsum[1]

% \subsection{Energy-minimizing paths}
% \label{app:metric-learning}
% \lipsum[1]

\subsection{Architecture Details}
\label{app:arch}

\begin{figure}
    \centering
    \includegraphics[width=0.8\linewidth]{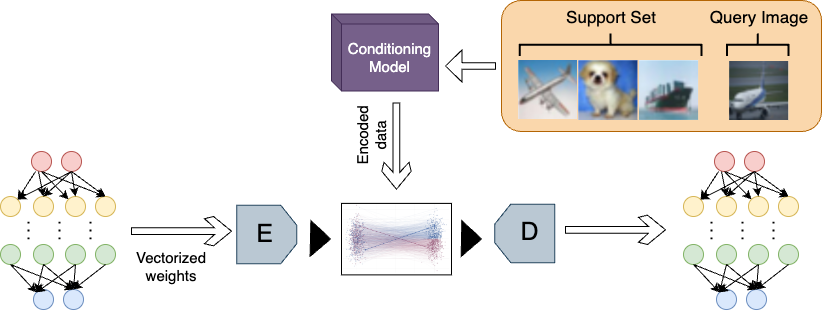}
    \caption{A schematic of the training process of \fsl{} for few-shot learning. Given a set of pre-trained target weights and a support set, we apply the conditioned flow model to pushforward a sample of the latent prior towards encoded target weights. The decoder is used during inference where we start from a sample of the latent prior and pushforward towards the target distribution with a trained vector field $v_\theta(\cdot, t; \vy)$ where $\vy$ is the support set embedding.}
    \label{fig:f2sl}
\end{figure}

Here, we expound on the architecture of \fsl{}. See Figure~\ref{fig:f2sl} for a schematic of the training and inference process.

\subsubsection{Variational Autoencoder}
The variational autoencoder follows the implementation of \citet{soroDiffusionbasedNeuralNetwork2024}. In particular, given a set of model weights $\{\mathcal{M}_i\}_{i = 1}^N$, we first flatten the weights to obtain vectors $\vw_i \in \R^{d_i}$. For the sake of uniformity, we always zero-pad vectors to $d = \max_i d_i$. Alternatively, we allow for layer-wise vectorization: set a chunk size $\ell$ which corresponds to the weight dimension of a network layer. Then, zero-pad $\vw_i$ to be a multiple of $\ell$, say $\tilde d$. This allows us to partition into $k$ equal length vectors $\vw_{i, k} \in \R^{\tilde d/k}$. Typically, larger models benefit from layer-wise vectorization.

Subsequently, we train a VAE to obtain an embedding of such vectors by optimizing the objective:
\begin{equation}
\label{eq:vae}
    L_{\text{VAE}}(\theta, \phi) := -\E_{\vz \sim q(\vz | \vw)} [\log p_\theta(\vw | \vz)  + \beta D_{KL}(q_\phi(\vz | \vw) || p(\vz))]
\end{equation}
where $\vw$ is the vectorized weights, $\vz$ is the embedding we are learning, and $p_\theta, q_\phi$ are the reconstruction and posterior distributions respectively. Moreover, we fix the prior $p(\vz)$ to be a $(0, 1)$-Gaussian and the weight is set to be $\beta = 10^{-2}$. For layer-wise vectorization, we simply change the input dimensions to match the chunk size. Upon decoding, we concatenate the chunks to re-form the weight vector.

\subsubsection{Graph Encoder}

Recently, \citet{kofinasGraphNeuralNetworks2024} proposed a neural graph encoder which incorporates the permutation invariance present in network weights. The method has two components: a graph constructor and the embedding model. The neural network is first represented as a graph where nodes represent the neurons within each network layer and edges represent neuronal connections. Importantly, node features correspond to bias parameters and edge features correspond to weight parameters. Subsequently, this is fed into an embedding model, such as a GNN, specifically PNA \citep{corso2020principalneighbourhoodaggregationgraph}, or a relational transformer \citep{diao2023relational}. For our use case, we outline weights-to-graphs conversion of MLPs, CNNs, and normalization layers. See \citet{kofinasGraphNeuralNetworks2024} for more details.

\paragraph{MLPs to graphs.} Let $\mathcal{G}(\mV, \mE)$ be a graph and let the vertex set $\mV \in \R^{n \times d_V}$ and the adjacency matrix $\mE \in \R^{n \times n \times d_E}$. Intuitively, if we have $n$ nodes in a graph, our vertex set is size $n$, and the adjacency matrix is $n \times n$. In our case, we also incorporate node and edge features, hence an extra dimension is added. Consider an $L$-layer MLP with weights $\{\mW^{\ell} \in \R^{d_{\ell} \times d_{\ell - 1}}\}_{\ell = 1}^L$ and biases $\{\vb^\ell \in \R^{d_\ell}\}_{\ell = 1}^L$. Since we have a node for each neuron, we have $n = \sum_{\ell = 0}^L d_\ell$, where $d_0$ is the input dimension. Now, let's use these to construct the vertex set $\mV$. Since each neuron has a corresponding bias term (except the input), $\mV = [\bf 0_{d_0} \; \vb^1 \; \dots \; \vb^L]^\top$. As for the adjacency matrix, consider the first $d_0$ rows: as this corresponds to the input layer, it's only connected to the first layer, i.e. only columns $d_0 + 1$ to $d_0 + d_1$ are possibly non-zero. And if we focus on row $i \in [d_0]$, what are its features? They must be $\mW^1_{:, i}$. Hence, \[
(\mE_{[0:d_0] \times [d_0 + 1, d_0 + d_1]})^\top= \mW^1,
\]
and elsewhere in $\mE_{[0:d_0]}$ is zero. In general, \[
(\mE_{[d_{i-1}:d_i] \times [d_{i-1} + 1, d_{i-1} + d_i]})^\top= \mW^i,
\]
and is zero everywhere else. In other words, the first off-diagonal blocks are precisely $\mW^i$, and $\mE$ is zero elsewhere. Finally, what are $d_E$ and $d_V$? This turns out to be problem-dependent. Sometimes, it helps to add useful node features, but if the only thing we are concerned about embedding is weight information, then each entry of $\mW^i$ and $\vb^i$ is simply a scalar, so $d_E = d_V = 1$.

\paragraph{Normalization layers to graphs.} Either BatchNorm or LayerNorm can be written as $\vy = \vm \odot \vx + \vb$, where $\vm, \vx, \vb, \vy \in \R^d$. The trick is to recast this as a linear layer: we can always write $\vy = \text{diag}(\vm) \vx + \vb.$ Hence, we ought to have $d$ nodes for $\vx$ and another $d$ nodes for $\vy$ where the nodes for $\vy$ have biases $\vb$. The two layers are then connected by weight matrix $\text{diag}(\vm)$ which only connects $x_i$ to $y_i$.

\paragraph{CNNs to graphs.} To simplify consider one convolutional layer between layers $\ell - 1$ and $\ell$, namely $\mW \in \R^{d_\ell \times d_{\ell - 1} \times w_\ell \times h_\ell}$ and $\vb \in \R^{d_\ell}.$ Intuitively, $d_{\ell - 1}$ is the number of input channels and $d_\ell$ the number of output channels. Due to the spatial dimension $w_\ell \times h_\ell$, we first flatten the last two layers. Now, we make use of the node and edge features: instead of scalar weights like in linear layers, our weights are vectors of size $w_\ell \times h_\ell$. However, the size may be different between layers, so we take $s = (\max_{\ell \in [L]} w_\ell, \max_{\ell \in [L]} h_\ell)$ and zero-pad our weight vectors as necessary before flattening. Hence, following the procedure in the MLP conversion, we form an adjacency matrix with vector features, i.e. $\mE \in \R^{n \times n \times d_E}$ where $d_E = w_{\max} h_{\max}$.

\subsubsection{Flow Model}

The neural network used for flow matching is the UNet from D2NWG \citep{soroDiffusionbasedNeuralNetwork2024}. The specific hyperparameters used for the CFM model varies between experiments, so we leave this discussion to~\ref{app:experimental-details}.

\begin{table}[t]
\caption{Model architectures and hyperparameters. Square brackets $[\cdot]$ indicates an interval of values. For instance, we often train until a loss plateau, hence the varying number of epochs.}
\label{tab:model_architectures}
\tablestyle{1.5pt}{1.0}
\centering
\begin{tabular}{lcc}
\toprule
Parameters & Model Retrieval & Few-Shot Learning \\
\midrule
\multicolumn{3}{c}{\textbf{Dataset Encoder (Frozen)}} \\
\midrule
Architecture & CAML & CAML\\
Latent Dimension & 1024 & 1024\\
\midrule
\multicolumn{3}{c}{\textbf{Weight Encoder}} \\
\midrule
Architecture & VAE & VAE\\
Latent Space Size & $4 \times 4 \times 4$ & $4 \times 8 \times 8$\\
Upsampling/Downsampling Layers & 5 & 4\\
Channel Multiplication (per Downsampling Layer) & (1, 1, 2, 2, 2) & (1,1,2,2)\\
ResNet Blocks (per Layer) & 2 & 2\\
KL-Divergence Weight & 0.01 & 1e-6\\
Optimizer & AdamW & AdamW\\
Learning Rate & $1 \times 10^{-3}$ & $1 \times 10^{-2}$\\
Weight Decay & $2 \times 10^{-6}$ & $2 \times 10^{-6}$\\
Batch Size & 32 & 128 \\
Training Epochs & 3000 & [100, 500]\\
\midrule
\multicolumn{3}{c}{\textbf{Conditional Flow Matching Model}} \\
\midrule
Timestep and Dataset Embedding Size & 128 & 128\\
Input Size & $4 \times 4 \times 4$ & $4 \times 8 \times 8$\\
Optimizer & AdamW & AdamW\\
Learning Rate & $1 \times 10^{-3}$ & $2 \times 10^{-4}$\\
Weight Decay & $2 \times 10^{-6}$ & $2 \times 10^{-6}$\\
Batch Size & 32 & 128 \\
Training Epochs & [3000, 10000] & [100, 500]\\
NFE & 100 & 100\\
\bottomrule
\end{tabular}

\vskip -0.10in
\end{table}

\subsection{Training Details}
\label{app:training}
Here, we present further training and experimental details.

\subsubsection{Pre-trained Model Acquisition}

\paragraph{Datasets and architectures.} We conduct experiments on a wide range of datasets, including CIFAR-10/100~\citep{krizhevsky2009cifar}, STL-10~\citep{coates2011stl10}, (Fashion/K)-MNIST~\citep{xiao2017fmnist, clanuwat2018kmnist}, USPS~\citep{hull1994usps}, and SVHN~\cite{netzer2011svhn}.  To evaluate our meta-model's ability to generate new subsets of network parameters, we conduct experiments on ResNet-18~\citep{He2015DeepRL}, ViT-Base~\citep{dosovitskiy2021animage}, ConvNeXt-Tiny~\citep{liu2022convnet2020s}, the latter two are sourced from timm \citet{rw2019timm}. As we shall detail below, small CNN architectures from a model zoo~\citep{schurholtModelZoosDataset2022} are also used for full-model generations.

\paragraph{Model pre-training.} For better control over the target distribution $p_1$, in experiments involving ResNet-18, ViT-Base, and ConvNeXt-Tiny, we pre-train these base models from scratch on their respective datasets. We follow \citet{wangNeuralNetworkDiffusion2024} and train the base models until their accuracy stabilizes. Further, we train the relevant subset (e.g. batch norm parameters for ResNet-18) for another 200 epochs, saving the weights at the end.

\paragraph{Model zoo.} The model zoo used for meta-training in the model retrieval setting, as described in Sec. \ref{sec:model_retrieval}, was sourced from \citep{schurholtModelZoosDataset2022}. As the base model, we employed their CNN-small architecture, which consists of three convolutional layers and contains either 2,464 or 2,864 parameters, depending on the number of input channels. For each dataset—MNIST, Fashion-MNIST, CIFAR-10, and STL10—100 sets of pre-trained weights were randomly selected from the model zoo using different seeds and fixed hyperparameters (referred to as "Seed" in their codebase). For the training of base models, we adopted the same hyperparameters as those used in \citep{schurholtModelZoosDataset2022} for all datasets, except KMNIST, which was not included in their model zoo. For KMNIST, we used the hyperparameters applied to MNIST, given the similarity between the two datasets.

\subsubsection{Variational Autoencoder Training}
The VAE was trained with the objective in \eqref{eq:vae}. Moreover, following p-diff \citep{wangNeuralNetworkDiffusion2024}, we add Gaussian noise to the input and latent vector, i.e. given noise factors $\sigma_{in}$ and $\sigma_{lat}$ with encoder $f_\phi$ and decoder $f_\theta$, we instead have \[
\vz = f_\phi(\vw + \xi_{in}), \; \hat \vw = f_\theta(\vz + \xi_{lat}) \quad \text{where} \quad  \xi_{in} \sim \mathcal{N}(0, \sigma_{in}^2\mI), \; \xi_{lat} \sim \mathcal{N}(0, \sigma_{lat}^2\mI).
\]

A new VAE is trained at every instantiation of the CFM model as architectures often differ in their input dimension for different experiments. However, they are trained with different objectives: the VAE is trained to minimize reconstruction loss. In all experiments, we fix $\sigma_{in} = 0.001$ and $\sigma_{lat} = 0.5$.

\begin{wraptable}{r}{0.5\textwidth}
\vspace{-3.5em}
\caption{Preprocessing and graph encoder hyperparameters.}
\label{tab:graph_enc}
\begin{tabular}{lcc}
\toprule
Parameters & & Values \\
\midrule
\multicolumn{3}{c}{\textbf{Dataset Preprocessing}} \\
\midrule
Input Channels & & 3 \\
Image shape & & (32, 32) \\
($w_{\max}, h_{\max})$ & & (7, 7) \\
Max. spatial res. &  & 49 \\
Max. \# hidden layers & & 5 \\
Flattening Method & & Repeat Nodes \\
Normalize & & False \\
Augmentation & & False \\
Linear as Conv. & & False \\
\midrule
\multicolumn{3}{c}{\textbf{Relational Transformer}} \\
\midrule
Embed dim. & & 64\\
Num. layers & & 4 \\
Num. heads & & 8 \\
Num. probe features & & 0\\
\bottomrule
\end{tabular}
\vspace{-5em}
\end{wraptable}

\subsubsection{Graph Encoder Training}

The graph encoder~\citep{kofinasGraphNeuralNetworks2024} was used for both the unconditional generation and fine-tuning on OOD experiments with the CNN-medium architecture from \citet{schurholtModelZoosDataset2022}. We restricted our tests to the relational transformer~\citep{diao2023relational} which was shown to perform better in the original paper~\citep{kofinasGraphNeuralNetworks2024}. See Table~\ref{tab:graph_enc} for the instantiation parameters.

\subsection{Experimental Details}
\label{app:experimental-details}
In this section, experimental hyperparameter details are given alongside estimates of computation time on an A100 GPU.

\subsubsection{Unconditional Generation}
Unconditional generation involves two stages: first is the training of base models. We choose a Resnet18, ViT-B, ConvNext-tiny, and medium-CNN for our base models and provide the training parameters in Table~\ref{tab:task-training}. Next, is the stage where we train either a AE-CFM or AE-DDPM, with the encoder being the same in both cases; the training parameters for this stage is provided in Table~\ref{tab:generative}. We found that in most cases, the autoencoder and CFM converge after 1000 epochs. In this case, we estimate training time to be between 2-3 hours when the VAE is used. When the graph encoder is used, of course the CFM training time remains the same, however, the encoder takes 6-8 hours of training. Validation requires less than 1 min. to run due to the small latent dimension.

\begin{table}[t]
\caption{Task Training}\label{tab:task-training}
\tablestyle{1.5pt}{1.0}
\centering
\begin{tabular}{lccc}
\toprule
Parameters & ResNet18 & ViT \& ConvNext & CNN \\
\midrule
Optimizer & SGD & AdamW & AdamW\\
Initial Training LR & 0.1 & $1\times 10^{-4}$ & $3\times 10^{-3}$\\
Training Scheduler & MultiStepLR & CosineAnnealingLR & CosineAnnealingLR \\
Layer Weights Saved & Last 2 BN layers & Last 2 BN layers & All layers\\
Initial Model Saving LR & $1.6\times 10^{-4}$ & $5\times 10^{-2}$ & $1\times 10^{-3}$\\
Model Saving Scheduler & None & CosineAnnealingLR & CosineAnnealingLR \\
Number of Models Saved & 200 & 200 & 200 \\
Num. of Weights per Model & 2048 & 3072 & [10565, 12743]\\
Training Epochs & 100 & 100 & 100\\
Batch Size & 64 & 128 & 128\\
\bottomrule
\end{tabular}

\vskip -0.10in
\end{table}

\subsubsection{Model Retrieval} 
\label{app:model-retrieval}
The first column of Table \ref{tab:model_architectures} shows the details of the model architectures and training configurations. For each dataset, we first generate its CAML embedding by (1) averaging the query image embeddings within each class to get class embeddings, (2) concatenating the class embeddings into one long vector, (3) passing the combined class embeddings through two linear layers to produce the final dataset embedding. Next, the dataset embedding is combined with the timestep embedding via a projection layer, and the resulting representation is used as input to the flow matching model. We estimate a training time of 2 hours for the VAE and 4 hours for the CFM to achieve our level of accuracy. Inference times remain the same as in \textit{unconditional generation}.

\subsubsubsection{A mini-Imagenet prior for CFM.} As mentioned in Sec.~\ref{sec:model_retrieval}, we attempted this experiment with priors from a pre-trained mini-Imagenet. There were a few technical hurdles with the implantation of these weights. First, for 1-channel datasets such as MNIST, the input weight shapes are smaller than those of the mini-Imagenet model. Second, the classification head of a mini-Imagenet model predicts a much greater number of classes than our test datasets. Our procedure is as follows: we train a small-CNN model~\citep{schurholtModelZoosDataset2022} on mini-Imagenet until its accuracy stabilizes. Next, we take the mean $\mu$ and standard deviation $\sigma$ of its classifier head. Using these statistics, we initialized the classifier heads of our base models as $\mathcal{N}(\mu, \sigma^2 \mI)$. For the rest of the base model, we pad to the prior's shape if necessary, and we implant the pre-trained weights directly before adding Gaussian $\mathcal{N}(\bf 0, \mI)$ noise. Since we flow in latent space, our last step is to apply the VAE to the weights we've constructed.

\begin{wraptable}{r}{0.5\textwidth}
\vspace{-2em}
\caption{Finetuning}\label{tab:fine-tuning-params}
\tablestyle{1.5pt}{1.0}
\centering
\begin{tabular}{lcc}
\toprule
Parameters & ResNet18 & CNN\\
\midrule
Optimizer & AdamW & AdamW \\
Num Epochs & [50,100] & [50, 100]\\
Initial LR & $1\times 10^{-2}$ & $1\times 10^{-5}$\\
Detach Value & 0.4 & 0.4\\
LR Scheduler & None & CosineAnnealingLR\\
Minimum LR & $1\times 10^{-2}$ & $5\times 10^{-7}$\\
\bottomrule
\end{tabular}
\end{wraptable}

\begin{figure}
    \centering
    \includegraphics[width=0.8\linewidth]{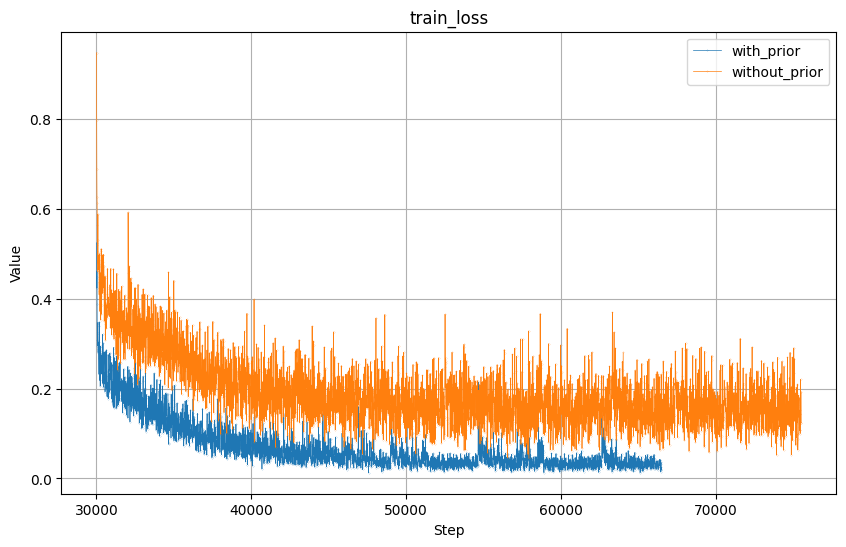}
    \caption{The training loss curve for the mini-Imagenet run from Table~\ref{tab:model-retrieval}.}
    \label{fig:training_curves_model_retrieval}
\end{figure}

Figure \ref{fig:training_curves_model_retrieval} shows the training curve with our mini-Imagenet prior in blue, and with a Gaussian 0-1 prior in orange. It is striking that the loss decreases much faster, but as seen in Table~\ref{tab:model-retrieval}, the test accuracies are quite poor. This points to an issue such as overfitting, which is likely caused by latent space capacity. Indeed, with our approach of constructing the prior, we invoke the VAE encoder twice: once to encode the prior and once more to encode the target weights. The target weights were those pre-trained on one of the four datasets in Table~\ref{tab:model-retrieval}, hence it's expected that the distribution is quite distinct from those pre-trained on mini-Imagenet. Due to the size of our latent space (64, as noted in Table~\ref{tab:model_architectures}), it may be insufficient to encode both distributions. Moreover, the loss objective for encoding the prior is not ideal. As the encoder is invoked in forward passes of the CFM, it only learns how to encode the prior such that CFM loss decreases, as opposed to a reconstruction objective. Hence, future work could look to modify encoder training so as to reconstruct both target weights and weights of the prior.

\subsubsection{Few-Shot Learning}
For the few-shot learning experiments, we adopted the same methodology for obtaining dataset embeddings and conditioning parameter generation as in the model retrieval experiment. Classifier heads were trained on 50,000 randomly sampled 5-way 1-shot subsets of the mini-ImageNet dataset \citep{vinyals2016matching}, and the resulting pre-trained weights were used as training data for the meta-model. The hyperparameter configurations for meta-training are provided in the second column of Table \ref{tab:model_architectures}. During evaluation, we measured the accuracies of the generated models on two out-of-distribution datasets, CIFAR-10 and STL-10. For each subset, we sampled 50 sets of weights and reported the average accuracy of the top three performing models. We estimate a training time of 4 hours for the VAE (until convergence) and 7 hours of CFM training until convergence. 

\begin{table}[t]
\caption{CFM/DDPM}\label{tab:generative}
\vspace{0.5em}
\tablestyle{1.5pt}{1.0}
\centering
\begin{tabular}{lcc}
\toprule
Parameters & AE CFM & AE DDPM\\
\midrule
Flow/Diffusion Optimizer & AdamW & AdamW \\
Flow/Diffusion LR & 0.001 & 0.001 \\
Autoencoder Optimizer & AdamW & AdamW\\
Num Inference Timesteps & 100 & [100, 1000]\\
Autoencoder LR & 0.001 & 0.001\\
Weight Initialization & Kaiming & Normal\\
Autoencoder Epochs & [1000, 30000] & [1000, 30000]\\
CFM/DDPM Epochs & [1000, 30000] & [1000, 30000]\\
Batch Size & [50, 200] & [50, 200]\\
\bottomrule
\end{tabular}
\vskip -0.10in
\end{table}

\subsubsection{Fine-tuning on Out-of-Distribution Data}
\label{app:fine-tuning}
We provide the hyperparameters of the fine-tuning experiment in Table~\ref{tab:fine-tuning-params}. The computational cost of this experiment mainly lie in the backpropagation step through the ODE solver used for CFM inference. We implemented a \verb|stopgrad| mechanism to trade off efficiency for accuracy, and this parameter was set to detach the computation graph at $t = 0.4$, therefore saving 40\% of backpropagation had we not used it. The total runtime depends on the dataset used, but we estimate a runtime of about 16 hours.

% \vspace{-.5em}
% \begin{wraptable}{r}{\textwidth}
% \caption{Fine-tuning on OOD data. Model-F are base models generated from fine-tuned meta-models, whereas Model-S are static.}
% \vspace{-0em}
% \label{tab:ft-resnet}
% \tablestyle{1.5pt}{1.0}
% \begin{center}
% \begin{adjustbox}{max width=0.9\textwidth}
% \begin{tabular}{l @{\hspace{10pt}} l @{\hspace{10pt}} c@{\hspace{10pt}}c@{\hspace{10pt}}c@{\hspace{10pt}}c}
% \toprule
% Base dataset & Type & CIFAR10 & CIFAR100 & STL10\\
% \midrule
% \raisebox{-0.4\normalbaselineskip}[0pt][0pt]{CIFAR10} &         Resnet18-S  & - & 68.80 & 64.20 \\
% &   Resnet18-F  & - & 70.96 & 72.87 \\
% \midrule
% \raisebox{-0.4\normalbaselineskip}[0pt][0pt]{CIFAR100}&         Resnet18-S  & 93.79 & -     & 56.46 \\
% &   Resnet18-F  & 94.07 & -     & 73.04\\
% \midrule
% \raisebox{-0.4\normalbaselineskip}[0pt][0pt]{STL10}   &         Resnet18-S  & 93.09 & 67.74 & - \\
% &   Resnet18-F  & 94.06 & 70.92 & - \\
% \bottomrule
% \end{tabular}
% \end{adjustbox}
% \vspace{-1em}
% \end{center}
% \end{wraptable}

\end{document}